\documentclass[letterpaper, 10 pt, conference]{ieeeconf}
\pdfoutput=1                   
\usepackage{graphics}
\usepackage{epsfig}
\usepackage{mathptmx}
\usepackage{times}
\usepackage{amsmath}
\usepackage{amssymb}
\usepackage{bm} 
\usepackage{cite}
\usepackage{url}
\usepackage{color}

\title{\LARGE \bf Design of a Parallel Elastic Actuator with a Continuously-Adjustable Equilibrium Position}

\author{Evangelos Chatziandreou, Chase W. Mathews, and David J. Braun 
\thanks{E. Chatziandreou, C. W. Mathews, and D. J. Braun are with the Advanced Robotics and Control Laboratory within the Center for Rehabilitation Engineering and Assistive Technology, Department of Mechanical Engineering, Vanderbilt University, Nashville, Tennessee 37235, USA. \newline
\indent This work was supported in part by a Seeding Success Grant provide by Vanderbilt University and a National Science Foundation CAREER Award (Grant No. 2144551). The authors gratefully acknowledge the support.}
\thanks{E-mail: {\tt\small evangelos.chatziandreou@vanderbilt.edu}}
\thanks{E-mail: {\tt\small chase.w.mathews@vanderbilt.edu}}
\thanks{E-mail: {\tt\small david.braun@vanderbilt.edu}}
}
\IEEEoverridecommandlockouts
\begin{document}
\maketitle
\begin{abstract}
In this paper, we present an adjustable-equilibrium parallel elastic actuator (AE-PEA). The actuator consists of a motor, an equilibrium adjusting mechanism, and a spring arranged into a cylindrical geometry, similar to a motor-gearbox assembly. The novel component of the actuator is the equilibrium adjusting mechanism which (i) does not require external energy to maintain the equilibrium position of the actuator even if the spring is deformed and (ii) enables equilibrium position control with low energy cost by rotating the spring while keeping it undeformed. Adjustable equilibrium parallel elastic actuators resolve the main limitation of parallel elastic actuators (PEAs) by enabling energy-efficient operation at different equilibrium positions, instead of being limited to energy-efficient operation at a single equilibrium position. We foresee the use of AE-PEAs in industrial robots, mobile robots, exoskeletons, and prostheses, where efficient oscillatory motion and gravity compensation at different positions are required.
\end{abstract}

\section{Introduction}
Parallel elastic actuators (PEAs) consist of a motor and a spring connected in parallel to the load \cite{Mettin2010}. In PEAs, the spring helps unload the motor by providing torque at net-zero energy cost, thus lowering the torque and the energy consumption of the motor \cite{Wang2011}. Parallel elastic actuators have shown promise in reducing motor torque in static tasks, such as lifting \cite{Toxiri2018}, and in dynamic tasks, such as robot locomotion \cite{Mazumdar2017,Liu2018} and human locomotion \cite{Li2020,Guo2022,Orekhov2022}.
However, PEAs are limited to a fixed equilibrium position; while the motor can efficiently generate oscillatory motion centered around the fixed equilibrum position of the actuator, it requires extra energy to displace the spring to a new equilibrium position or to efficiently generate oscillatory motion away from the equilibrium position. This feature limits the benefit of PEAs in static gravity compensation and oscillatory pick-and-place tasks, which may require holding loads at different positions.

\begin{figure} [t!]
	\centering
	\includegraphics[width=\columnwidth]{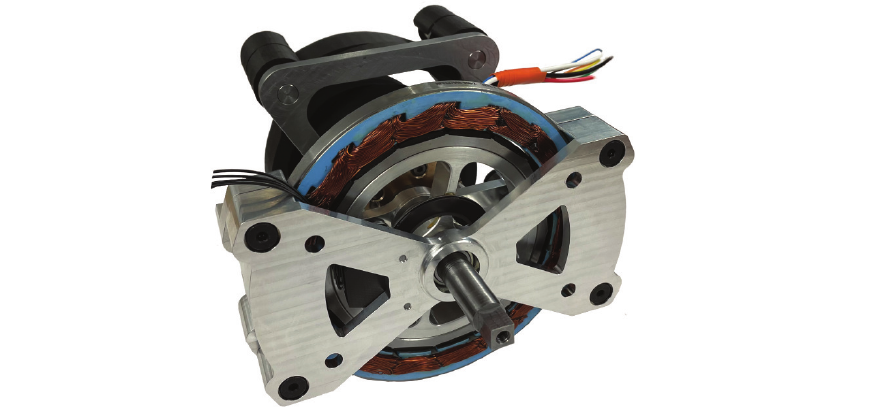}
	\caption{Prototype adjustable-equilibrium parallel elastic actuator.}
	\label{IntroImage}
\end{figure}

\begin{figure*}[ht]
	\centering
	\includegraphics[width=\textwidth]{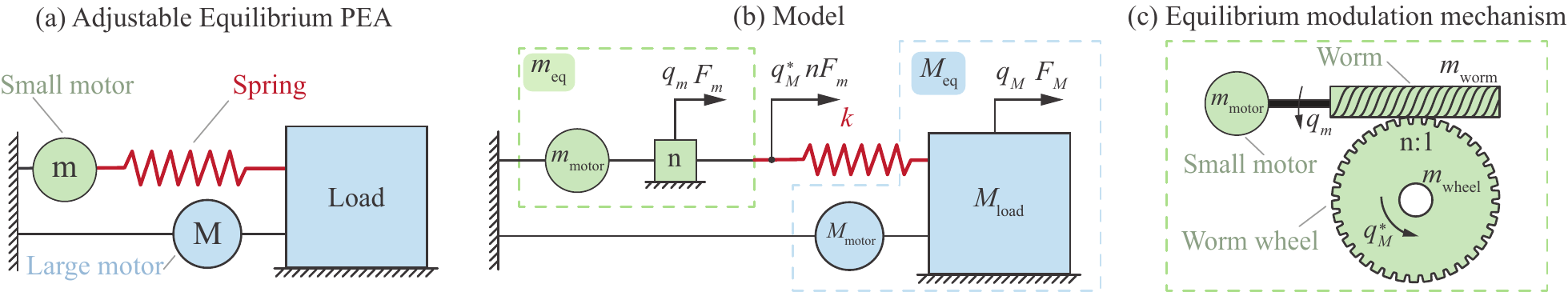}
	\caption{(a) Adjustable Equilibrium Parallel Elastic Actuator. (b) Model of the AE-PEA. (c) Model of the equilibrium adjusting mechanism.}
	\label{Math_model}
\end{figure*}

In order to alleviate the limitation of PEAs due to their fixed equilibrium position, clutched parallel elastic actuators (CPEAs) have been previously developed. 
In CPEAs the spring is disengaged from the motor and the load through the use of locking devices, such as electromagnetic clutches \cite{Haeufle2012} or planetary differentials and brake pads \cite{Plooij2016}. To change the equilibrium position of the actuator, CPEAs disengage the spring when it is un-deflected or locked, use the motor to drive the load while the spring is disengaged, and subsequently, re-engage the spring at a new position.
Adequate control of CPEAs avoids disengaging or re-engaging the spring when deflected, as the former leads to energy dissipation, while the latter introduces extra energy as the spring recoils into the system. 

An alternative way to change the static equilibrium position of PEAs is to modify the stiffness of the actuator. For example, changing the stiffness of the spring can vary the resting position of the actuator when subject to a static load \cite{Dorsser2008,Kim2022}. 
This can be done using parallel variable stiffness actuators (PVSAs) recently introduced in \cite{Mathews2021}.
However, the equilibrium position of PVSAs is limited by the deformation range of the spring.
The stiffness and equilibrium position of the actuator can be simultaneously adjusted by using multiple clutched springs \cite{Plooij2017,Krimsky2020} or by using a floating spring mechanism \cite{Kim2021}.
These actuators allow discrete adjustments of the stiffness and equilibrium position as a function of the locking and unlocking of the springs.

In this paper, we introduce an adjustable-equilibrium parallel elastic actuator (AE-PEA), which consists of a direct-drive motor arranged in parallel with a spring that has a continously-adjustable equilibrium position, see Fig.~\ref{IntroImage}.
The device features a 3D printed carbon-fiber reinforced torsion spring, one end of which is coupled to the main direct-drive motor while the other end is attached to a worm drive. In order to change the equilibrium position of the actuator, a small but fast motor is used to turn the worm drive simultaneously with the rotation of the direct-drive motor and the load to keep the spring undeformed.
In this way, the actuator can set the equilibrium position of the actuator to any position, irrespective of the deformation range of the spring. Also, because the worm drive self-locks under load, the actuator requires no energy to maintain a given equilibrium position, and any unexpected external disturbance automatically re-engages the spring, rendering an energetically passive response of the actuator independent of the control of the spring.
In summary, AE-PEAs extend the benefits of PEAs from a single equilibrium position to any arbitrary equilibrium position.  

\section{Model} \label{model}
A conceptual model of the adjustable-equilibrium parallel elastic actuator is shown in Fig.~\ref{Math_model}a. The AE-PEA consists of three main components: (i) the spring, (ii) the large motor attached to the load in parallel with the spring, and (iii) the equilibrium adjusting mechanism driven by the small motor, shown in Fig.~\ref{Math_model}b-c. In the reminder of this section, we present the mathematical model of the AE-PEA.

\subsection{Model}
Neglecting the internal friction in the actuator, the dynamics of the load and the equilibrium position adjusting system during forward drive is given by the following two equations:
\begin{equation}
	M_{\text{eq}}\Ddot{q}_M=F_M+F_{\text{spring,}M},
	\label{EOM-Motor-definition}
\end{equation}
\begin{equation}
	m_{\text{eq}}\Ddot{q}_m=F_m+F_{\text{spring,}m},
	\label{EOM-motor-definition}
\end{equation}
where $M_{\text{eq}}$ is the equivalent mass of the actuator, $\Ddot{q}_M$ is the acceleration of the main motor and the load, $F_M$ is the force provided by the main motor, $F_{\text{spring,}M}$ is the force that the spring applies to the load, $m_{\text{eq}}$ is the equivalent mass of the equilibrium modulation mechanism reflected to the shaft of the small motor, $\Ddot{q}_m$ is the acceleration of the small motor, $F_m$ is the force provided by the small motor, while $F_{\text{spring,}m}$ the force that the spring exerts to the small motor through the equilibrium adjusting mechanism during forward drive.

The equivalent masses, $M_{\text{eq}}$ and $m_{\text{eq}}$, are defined by
\begin{equation}\label{Meq}
M_{\text{eq}}=M_{\text{motor}}+M_{\text{load}},
\end{equation}
\begin{equation}
	m_{\text{eq}}=m_{\text{motor}}+m_{\text{worm}}+\frac{1}{n^2}m_{\text{worm wheel},}
	\label{meq}
\end{equation}
where $n$ is the transmission ratio between the worm and the worm wheel shown in Fig.~\ref{Math_model}c. 

The forces applied by the spring to the load and the small motor are given by
\begin{equation}
	F_{\text{spring,}M}=-nF_{\text{spring,}m} = -k\Delta l,
	\label{Fspring,M}
\end{equation}
where $k$ is the stiffness of the spring, while $\Delta l$ is the deformation of the spring; the latter is defined by
\begin{equation}
	\label{springdeflection}
	\Delta l=q_M-\frac{1}{n}q_m.
\end{equation}

The model presented in (\ref{EOM-Motor-definition}) and (\ref{EOM-motor-definition}) does not capture the self-locking feature of the worm wheel drive, which implies that the small motor can move the end point of the spring but the spring cannot move the small motor, regardless of how much force it exerts to the worm wheel. In order to capture the essential behavior of the stiffness adjusting mechanism during backward drive, we neglect the transient inertial dynamics, and note that the following self-locking condition must hold, 
\begin{equation}
	F_m = 0 \Rightarrow \ddot{q}_m=0,\; \dot{q}_m=0.
	\label{nonbackdrivability condition}
\end{equation}

Although (\ref{EOM-Motor-definition}), (\ref{EOM-motor-definition}), and (\ref{nonbackdrivability condition}) provide one of the simplest models of the AE-PEA, this model neglects the internal load-dependent friction due to the worm drive \cite{Dohring1993}. 
We make this assumption because, as will be subsequently shown, the main purpose of the small equilibrium adjusting motor is to nullify the spring force, and thereby, the load-dependent friction in the actuator. 
This allows the simplified model to be suitable for the explanation of the working princple of the actuator.

\section{Working principle} \label{concept}
\begin{figure}[t]
	\centering
	\includegraphics[width=\columnwidth]{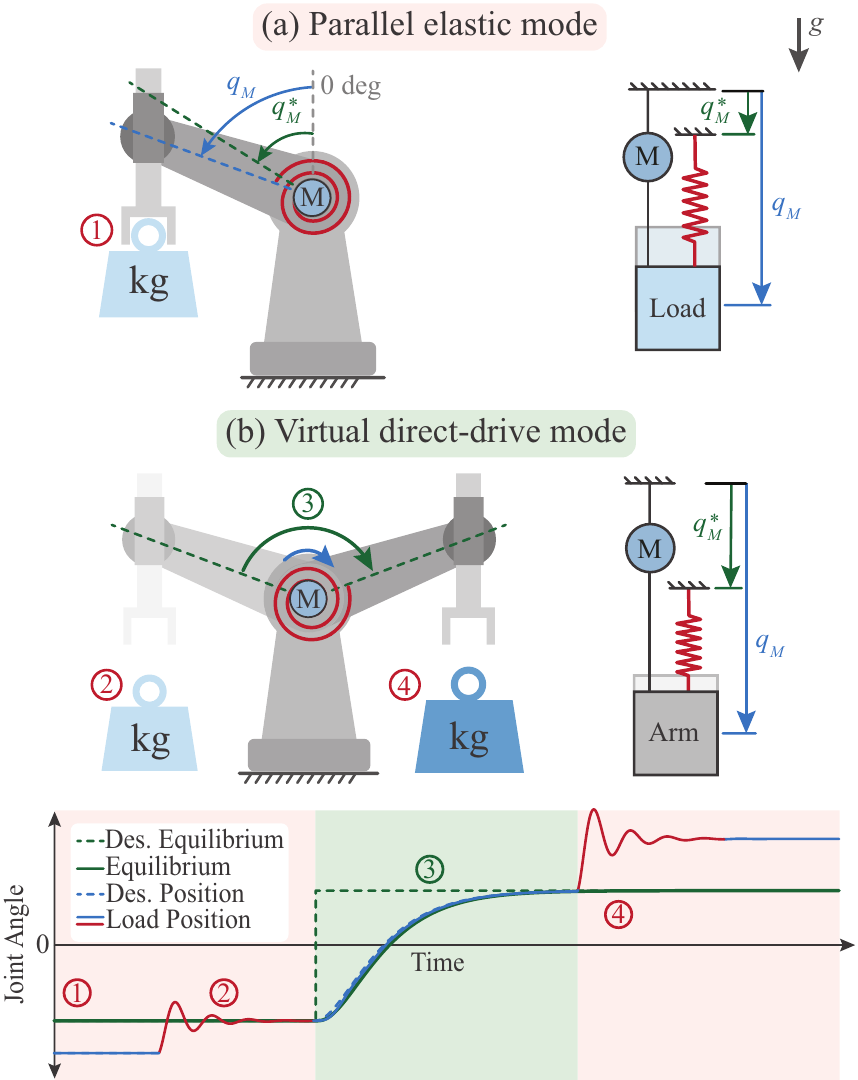}
	\caption{Example demonstrating the operation modes of a AE-PEA. A pick-and-place task is illustrated, where the robot holds and releases a mass at one equilibrium position, changes the equilibrium position, and picks up a mass at a different equilibrium position.
		\label{Task}}
\end{figure}

The working principle of the AE-PEA can be analyzed by considering two distinct operation modes of the actuator. 
In the first mode, the small motor is left uncontrolled $F_m=0$, and the AE-PEA behaves as a conventional PEA; we refer to this mode as parallel elastic mode, see Fig.~\ref{Task}a. 
In the second mode, the small motor is controlled to keep the spring uncompressed, and the AE-PEA behaves as a direct-drive actuator; we refer to this mode as virtual direct-drive mode, see Fig.~\ref{Task}b. In the second mode of operation, the large motor drives the load and controls the equilibrium position of the actuator.
In the remainder of this section we discuss the aforementioned two operation modes of the AE-PEA separately. 

\subsection{Parallel elastic mode}\label{S1}
If the the small motor is left uncontrolled, 
\begin{equation}
	\label{sl}
	F_m=0\Rightarrow \ddot{q}_m=\dot{q}_m=0\;\;\; \text{and}\;\;\; q_m=q_m^*,  
\end{equation}
then the AE-PEA behaves as a conventional PEA,
\begin{equation}
	M_{\text{eq}}\Ddot{q}_M=F_M-k( q_M - q^*_M),
	\label{EOM-Motor-definition0}
\end{equation}
with equilibrium position $q_M^*$ defined by the position of the small motor:
\begin{equation}
q_M^* = \frac{1}{n}q^*_m.  
\end{equation}
  
Because the worm wheel drive is self-locking, no force is required to operate the AE-PEA as a conventional PEA. Therefore, no energy is required to operate the AE-PEA in the parallel elastic mode, irrespective of the equilibrium position $q_M^*$ of the actuator. Also, because the small motor does not move, the forces between the worm and the worm wheel do not dissipate energy, and equation (\ref{EOM-Motor-definition0}) is an accurate representation of the actuator dynamics.

\subsection{Virtual direct-drive mode} \label{S2}
In order to adjust the equilibrium position of the actuator, we use the small motor to keep the spring undeformed by setting the desired spring length to zero,
\begin{equation}
	\label{desired1}
\Delta l_{\text{desired}} = q_M- \frac{1}{n}q_m = q_M - q^*_M = 0, 
\end{equation}
because an undeformed spring will ensure that the spring force is zero
\begin{equation}
F_{\text{spring,}M}=0 \;\;\; \text{and} \;\;\; F_{\text{spring,}m} =0. 
\end{equation}
Under such condition,
any potential load-dependent internal friction between the worm and the worm wheel will be negligible, and the actuator will behave as if it did not include the parallel spring,
\begin{equation}
	M_{\text{eq}}\Ddot{q}_M=F_M \;\;\; \text{and} \;\;\;  m_{\text{eq}}\Ddot{q}_m=F_m.
	\label{EOM-motor-definition1}
\end{equation}
\begin{figure*}[ht]
	\centering
	\includegraphics[width=\textwidth]{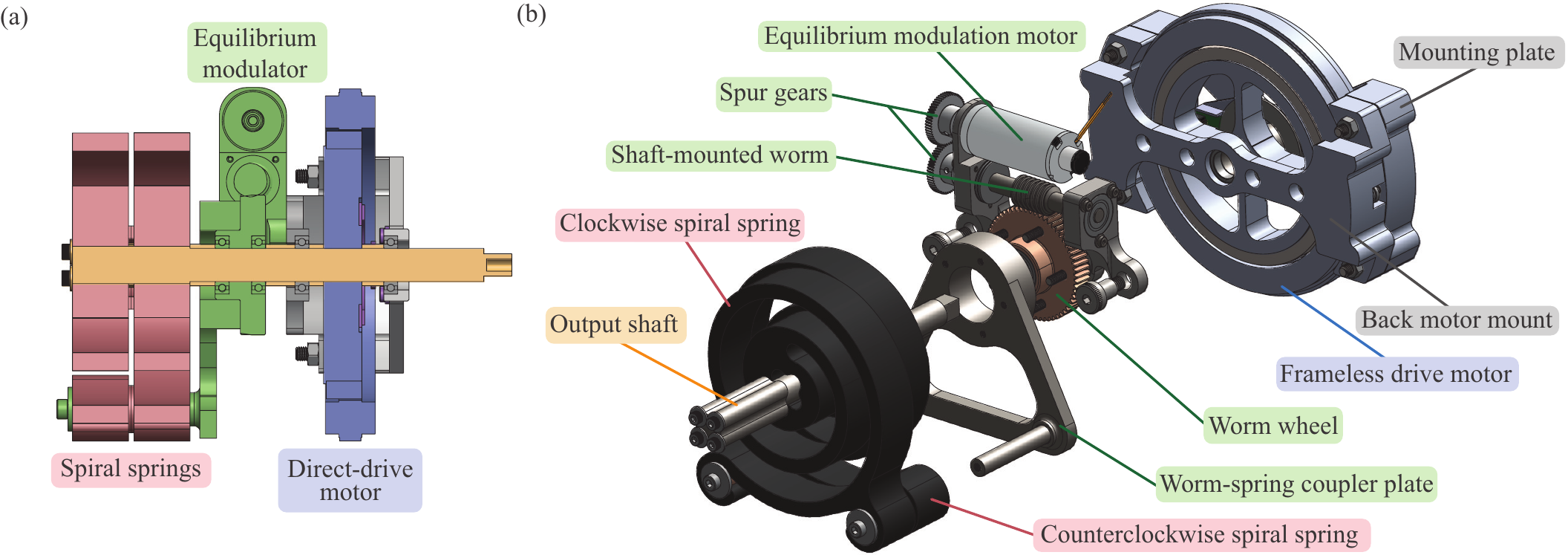}
	\caption{Prototype of the Adjustable-Equilibrium Parallel Elastic Actuator. (a) Section view of the AE-PEA showing the three subsystems; the motor (blue), the equilibrium adjusting mechanism (green), and the spring (red). (b) Exploded view of the AE-PEA. The actuator fits into a cylindrical shape of diameter $158$~mm and height $120$~mm. The mass of the actuator is $1.9$~kg.}
	\label{Section-Exploded views}
\end{figure*}

Based on (\ref{EOM-motor-definition1}), we conjecture that keeping the spring undeformed with the small motor will enable equilibrium position adjustment with a small amount of actuation energy, irrespective of the control implemented by the large motor to move the load.
 
In order to provide theoretical support for our conjecture, we will now derive a feedback control law for the motor force $F_m$ to implement $\Delta l \approx \Delta l_\text{desired} = 0$ (\ref{desired1}). For this purpose, we replace (\ref{desired1}) with an asymptotically stable desired dynamics
\begin{equation}
	\label{desired2}
\Delta \ddot {l} + 2\alpha \Delta \dot {l} + \alpha^2 \Delta l = 0,  
\end{equation}
where $\alpha >0$ defines how fast $\Delta l$ approaches $\Delta l_{\text{desired}}=0$ (larger $\alpha$ leads to faster response).
Subsequently, we substitute (\ref{EOM-Motor-definition}), (\ref{EOM-motor-definition}) and (\ref{springdeflection}) into (\ref{desired2}), to obtain the control force
\begin{equation}
	\label{control}
F_m=n\frac{m_\text{eq}}{M_\text{eq}}F_M \underbrace{- (1+n^2\frac{m_\text{eq}}{M_\text{eq}})F_{\text{spring,}m} + nm_\text{eq} (2\alpha \Delta \dot{l} + \alpha^2\Delta l),}_{\Delta l\neq 0\;\; \text{and}\;\; \Delta \dot{l}\neq 0}
\end{equation}
which is required by the small motor to ensure the length of the spring behaves according to the asymptotically stable desired dynamics (\ref{desired2}).

Assuming that the small motor can be reasonably well implemented as in (\ref{control}), we find
\begin{equation}
\Delta l \approx 0\;\; \text{and}\;\; \Delta \dot{l} \approx 0 \Rightarrow 	F_m \approx n\frac{m_\text{eq}}{M_\text{eq}}F_M.
\end{equation}

At this point, we see that in order to make the force of the small motor negligible compared to the force of the main motor, the design of the AE-PEA must satisfy the following condition:
\begin{equation}
	\label{condition}
n\frac{m_\text{eq}}{M_\text{eq}} \ll 1.
\end{equation}

Condition (\ref{condition}) can be satisfied in most practical applications by choosing a small mass ratio: $m_\text{eq}/M_\text{eq}\ll 1$. Nevertheless, condition (\ref{condition}) shows that excessive gearing is not desirable, not only because it limits the speed at which the small motor can implement (\ref{desired1}), but also because it increases the force required by the small motor to implement (\ref{desired1}).

Finally, we note that in order to change the equilibrium position of the AE-PEA, the small motor only ensures that the spring remains undeformed, regardless of the motion of the load controlled by the large motor. Therefore, any position of the large motor is a potential new equilibrium position of the acuator, as shown by (\ref{desired1}). 
To convert the position of the large motor into a new equilibrium position, the actuator can simply switch from virtual direct-drive mode (\ref{condition}) to parallel elastic mode (\ref{sl}) by stopping the small motor when $q_M= q^*_{M,\text{desired}}$.

\section{Design} \label{design}
In this section, we present a prototype of the actuator shown in Fig.~\ref{IntroImage}. The actuator was designed to fit into a compact cylindrical geometry, similar to standard motor-gearbox assemblies, and to have a modular structure with replaceable motors and springs, similar to configurable gearboxes. Figure~\ref{Section-Exploded views} shows the three main components of the prototype; (i) the direct-drive motor (blue), (ii) the custom-made composite spiral spring (red), and (iii) the equilibrium adjustment mechanism (green). In the following, we present these three component in detail.

\subsection{Direct-Drive Motor}
We use an Allied Motion MF0127008 brushless frameless motor shown in Fig.~\ref{Section-Exploded views}a (blue). The motor has a diameter of $127$~mm and a height of $17.8$~mm with a mass of $0.50$~kg. The continuous torque rating of the motor is $1.6$~Nm. The motor is mounted inside an aluminum housing, while the rotor is press-fit into the main shaft of the actuator, as shown in Fig.~\ref{Section-Exploded views}. A magnetic encoder (AMS AS5304A) which is embedded in the aluminum housing, and a sensor magnet (RLS MR045) which is embedded into the rotor, are used to measure the angular position of the motor shaft.

\subsection{Composite Spring}
The actuator uses two identical composite springs, shown in Fig.~\ref{Section-Exploded views}a (red). 
The springs were printed using the Markforged Mark Two printer. The 3D printing of the springs allows the diameter, width, and thickness of the springs to be modified for tasks that require different maximal deflection angles and different spring stiffness values. The springs used in the prototype were made from Onyx (nylon filled with chopped carbon fiber) reinforced with continuous strands of carbon fiber \cite{Mathews2021}. The springs have an estimated deflection range of $75$~deg and torsional stiffness of $21$~Nm/rad.

In the prototype, the inner diameter of the springs is coupled with the main shaft, while the outer diameter of the springs is connected to the worm wheel though a worm-spring connector plate, as shown in Fig.~\ref{Section-Exploded views}b. This arrangement allows the worm wheel to rotate the spring in unison with the rotation of the main shaft without deflecting the spring. 

Finally, we note that torsional springs may not have symmetric torque-angle characteristics in extension and compression. In order to mitigate the asymmetry in the torque-angle characteristic, the two torsional springs are mounted in opposite directions such that one is in extension while the other is in compression at all times. 
\begin{figure*}[ht]
	\centering
	\includegraphics[width=\textwidth]{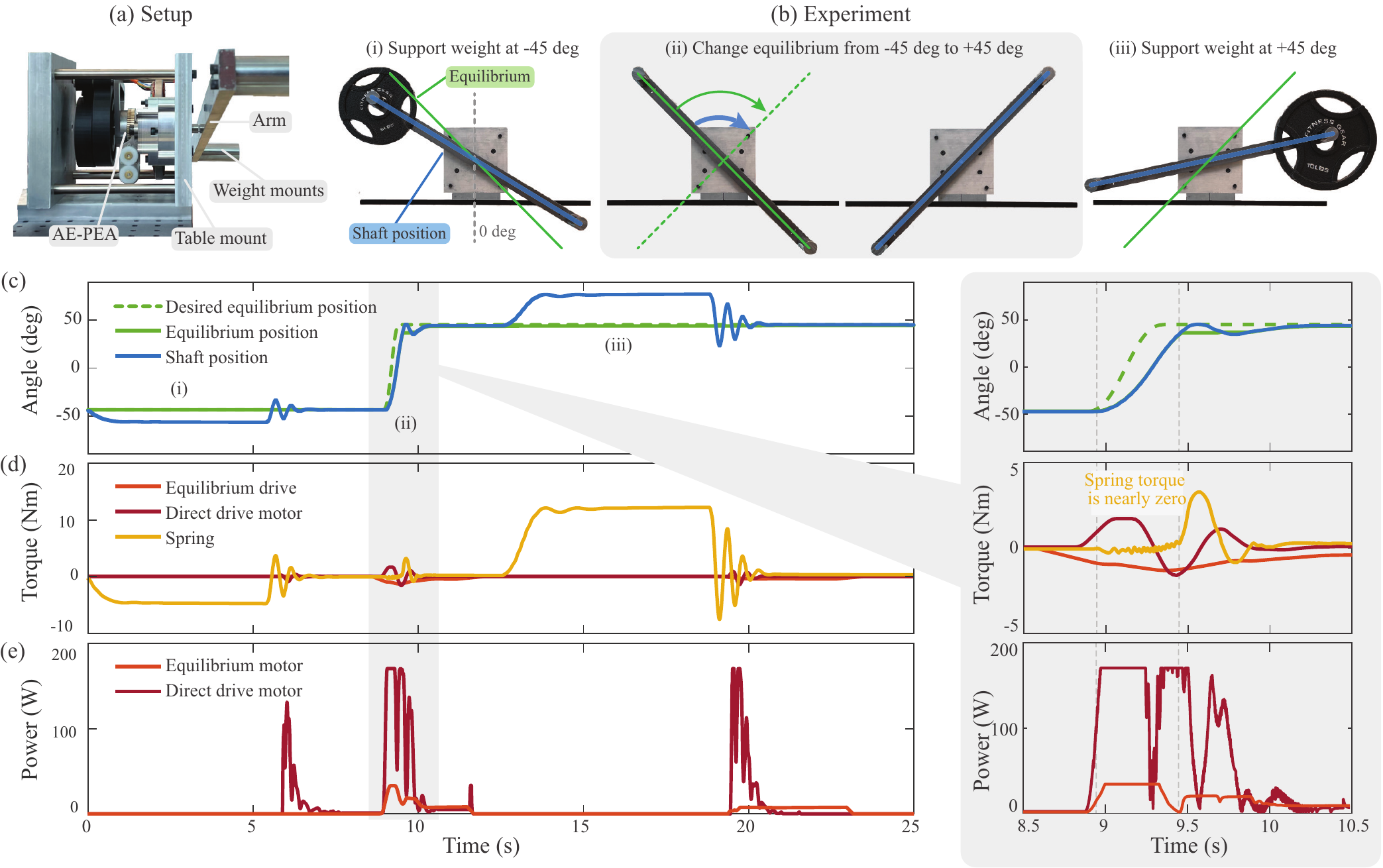}
	\caption{(a) Experimental setup. (b) Frame sequence that shows the three phase experiment. (c) Position of the equilibrium and the load. (d) Torque of the drive motor, equilibrium motor, and the spring. (e) Power consumption of the drive motor and the equilibrium position adjusting motor.
		\label{Experiment}}
\end{figure*}
\subsection{Equilibrium Adjustment Mechanism}
The equilibrium-changing mechanism in Fig.~\ref{Section-Exploded views}a (green), consists of a worm, worm gear, small motor, and a worm-spring connector plate, shown in Fig.~\ref{Section-Exploded views}b.
The worm and the worm wheel were chosen with a speed ratio of 60:1 such that self-locking occurs under load \cite{Kong2012}.
The worm shaft was aligned to the motor housing with a set of needle thrust bearings and ball bearings. 
The inside of the worm wheel floats about the main shaft on two ball bearings, while the outside of the worm wheel is bolted to the worm-spring connector plate, which couples the worm to the outer edge of the spiral springs.

We use a Maxon DCX22L motor to drive the worm. The motor has a nominal rotational speed of 9020~rpm and a nominal torque of 30.3~mNm with a small form factor (diameter of $22$~mm, length of $49$~mm, and a mass of $104$~g). Through the 60:1 gear ratio, the small motor can produce considerable torque to accelerate, decelerate, and rotate the spring at speeds up to $150$~rpm. We use an encoder on the small motor (Maxon ENX10) to measure the rotation of the worm and estimate the equilibrium position of the actuator.

\section{Evaluation} \label{experiments}
To verify the predicted advantages of the adjustable-equilibrium parallel elastic actuator, we conducted a simple experiment in which the actuator emulates a robot manipulator. We tested the actuator (i) holding a weight at no energy cost for an arbitrary desired equilibrium position and (ii) rapidly changing from one equilibrium position to a new equilibrium position.
These two tasks demonstrate the ability of the AE-PEA to switch between (i) the parallel elastic mode (Section~\ref{S1}), optimal for zero energy-cost static load bearing and passive disturbance rejection, and (ii) the virtual direct-drive mode (Section~\ref{S2}), optimal for quickly adjusting the equilibrium position of the actuator at low energy cost.

\subsection{Experimental Setup and Protocol}
Figure~\ref{Experiment}a-b show the experimental setup where the AE-PEA is mounted to a desk and is coupled to a steel bar of length $0.61$~m and mass $1.9$~kg. During the experiment, we attached two different masses ($2.3$~kg and $4.5$~kg) to the bar to emulate different payloads.

The position of the load $q_M$ was measured using the rotary encoder on the drive motor.
The equilibrium position of the actuator was computed by dividing the angle of the small motor (measured by the encoder mounted on the small motor) with the transmission ratio of the worm-drive $q_M^*=\frac{1}{n}q_m$ .
The torque of the large motor and the small motor were estimated using the commanded current of the motors and the corresponding torque constants, $\tau_M=k_MI_M$ and $\tau_m=k_mI_m$. 
The torque provided by the spring was estimated by multiplying the deflection of the spring $\Delta l= q_M - q_M^*$ with the experimentally determined stiffness of the actuator $k=21$~Nm/rad.
The electrical power consumption of the motors was estimated by multiplying the commanded current of each motor by the supply voltage, $p_M=I_M V$ and $p_m=I_m V$.

The experiment was performed in three phases.

(i) In the first phase, the actuator was set to a -$45$~deg equilibrium angle, and subsequently, it was first loaded and then unloaded using a $2.3$~kg weight (Fig.~\ref{Experiment}b-i).

(ii) In the second phase, the actuator changed the equilibrium angle from -$45$~deg to +$45$~deg (Fig.~\ref{Experiment}b-ii). 

(iii) In the third phase, the experiment in the first phase was repeated with a larger $4.5$~kg weight (Fig.~\ref{Experiment}b-iii).
\subsection{Results}
Figure~\ref{Experiment}c-e shows the experimental results.
Figure~\ref{Experiment}c shows the desired position of the equilibrium (dashed green line), the measured position of the equilibrium (green line), and the position of the motor shaft (blue line).
We observe that the actuator was able to hold the load during the parallel elastic operation mode, with no change to the equilibrium position. During the virtual direct-drive mode, the equilibrium position and motor position moved together from the starting equilibrium position at -$45$~deg to the new equilibrium position at +$45$~deg in around $1$~second.

Figure~\ref{Experiment}d shows the torque of the equilibrium drive (orange line), the direct-drive motor (red line), and the estimated torque of the spring (yellow line).
We observe that the motor and the equilibrium drive each supplied zero torque during the holding phase, while the spring provided an estimated torque of $4.7$~Nm and $12$~Nm to support the two different loads, respectively. During the virtual direct-drive phase, the direct-drive motor and the equilibrium drive provided a maximum torque of $1.6$~Nm and $1.3$~Nm, respectively, while the spring torque was kept below $0.3$~Nm.

Figure~\ref{Experiment}e shows the power consumption of the equilibrium drive (orange) and the direct-drive motor (red). During equilibrium change, the small motor and the direct-drive motor consumed an average power of $17$~W and $80$~W, respectively.
Both motors consumed $0$~W of power during the holding periods, except when the actuator was unloaded and the motors attempted to dampen the oscillations. If no spring were used, the direct-drive motor would require over $400$~W and $2500$~W of power to hold each load, respectively.

\section{Discussion and Conclusion} \label{discussion}
In this paper, we presented the design of a parallel elastic actuator in which the equilibrium position can be adjusted using a small motor. We showed the working principle of the AE-PEA, and experimentally demonstrated the ability of the actuator to (i) provide weight-bearing support at an arbitrary equilibrium position with zero energy cost, and (ii) adjust the equilibrium position without using the motors to work against the spring. These features extend the benefits of conventional parallel elastic actuation, where the spring is centered around a fixed equilibrium position, to parallel elastic actuation where the spring is centered at any desired equilibrium position.

The arrangement of the large motor in AE-PEAs differs from the arrangement of the large motor in series elastic actuators (SEAs) and series variable-stiffness actuators (SVSAs), as in both cases, the spring is in series with the large motor. 
Owing to this difference, AE-PEAs can provide high-fidelity force control compared to SEAs and SVSAs.
We note that AE-PEAs can be seen as a SEA driven by the small motor which is connected in parallel to the large direct-drive motor.
We note that the small motor limits the position control fidelity of the AE-PEA, but does not affect the force control fidelity of the actuator.

When comparing AE-PEAs to SVSAs, we observe that AE-PEAs use the same number of motors as a SVSA \cite{Braun2019b}, however, instead of using a small motor to change the stiffness of the actuator and a large motor to change the equilibrium position of the actuator, AE-PEAs use a small motor to change the equilibrium position of the actuator and a large motor to provide high fidelity force control.
Although AE-PEAs can make the stiffness of the actuator differ from the stiffness of the parallel spring (using feedback control) they cannot change the passive stiffness of the actuator.

In future works, we aim to combine AE-PEAs with the concept of parallel variable stiffness actuators, recently introduced by the authors, to alleviate the aforementioned limitations of the proposed concept of the AE-PEA.

\bibliographystyle{ieeetr}
\bibliography{bibliographyentries}{}

\begin{thebibliography}{10}

\bibitem{Mettin2010}
U.~Mettin, P.~X.~L. Hera, L.~B. Freidovich, and A.~S. Shiriaev, ``Parallel
  elastic actuators as a control tool for preplanned trajectories of
  underactuated mechanical systems,'' {\em The International Journal of
  Robotics Research}, vol.~29, no.~9, pp.~1186--1198, 2010.

\bibitem{Wang2011}
S.~Wang, W.~van Dijk, and H.~van~der Kooij, ``Spring uses in exoskeleton
  actuation design,'' in {\em IEEE International Conference on Rehabilitation
  Robotics}, (Zurich, Switzerland), pp.~1--6, June-July 2011.

\bibitem{Toxiri2018}
S.~Toxiri, A.~Calanca, J.~Ortiz, P.~Fiorini, and D.~G. Caldwell, ``A
  parallel-elastic actuator for a torque-controlled back-support exoskeleton,''
  {\em IEEE Robotics and Automation Letters}, vol.~3, no.~1, pp.~492--499,
  2018.

\bibitem{Mazumdar2017}
A.~Mazumdar, S.~J. Spencer, C.~Hobart, J.~Salton, M.~Quigley, T.~Wu,
  S.~Bertrand, J.~Pratt, and S.~P. Buerger, ``{Parallel elastic elements
  improve energy efficiency on the STEPPR bipedal walking robot},'' {\em
  IEEE/ASME Transactions on Mechatronics}, vol.~22, no.~2, pp.~898--908, 2017.

\bibitem{Liu2018}
X.~Liu, A.~Rossi, and I.~Poulakakis, ``A switchable parallel elastic actuator
  and its application to leg design for running robots,'' {\em IEEE/ASME
  Transactions on Mechatronics}, vol.~23, no.~6, pp.~2681--2692, 2018.

\bibitem{Li2020}
Y.~Li, X.~Guan, Z.~Li, Z.~Tang, B.~Penzlin, Z.~Yang, S.~Leonhardt, and L.~Ji,
  ``Analysis, design, and preliminary evaluation of a parallel elastic actuator
  for power-efficient walking assistance,'' {\em IEEE Access}, vol.~8,
  pp.~88060--88075, 2020.

\bibitem{Guo2022}
S.~Guo, R.~D. Gregg, and E.~Bolívar-Nieto, ``Convex optimization for spring
  design of parallel elastic actuators,'' in {\em 2022 American Control
  Conference (ACC)}, (Atlanta, GA, USA), pp.~3688--3694, June 2022.

\bibitem{Orekhov2022}
G.~Orekhov and Z.~F. Lerner, ``Design and electromechanical performance
  evaluation of a powered parallel-elastic ankle exoskeleton,'' {\em IEEE
  Robotics and Automation Letters}, vol.~7, no.~3, pp.~8092--8099, 2022.

\bibitem{Haeufle2012}
D.~F.~B. Haeufle, M.~D. Taylor, S.~Schmitt, and H.~Geyer, ``A clutched parallel
  elastic actuator concept: Towards energy efficient powered legs in
  prosthetics and robotics,'' in {\em IEEE RAS EMBS International Conference on
  Biomedical Robotics and Biomechatronics}, (Rome, Italy), pp.~1614--1619, June
  2012.

\bibitem{Plooij2016}
M.~Plooij, M.~Wisse, and H.~Vallery, ``{Reducing the energy consumption of
  robots using the bidirectional clutched parallel elastic actuator},'' {\em
  IEEE Transactions on Robotics}, vol.~32, no.~6, pp.~1512--1523, 2016.

\bibitem{Dorsser2008}
W.~D. van Dorsser, R.~Barents, B.~M. Wisse, M.~Schenk, and J.~L. Herder,
  ``Energy-free adjustment of gravity equilibrators by adjusting the spring
  stiffness,'' {\em Proceedings of the Institution of Mechanical Engineers,
  Part C: Journal of Mechanical Engineering Science}, vol.~222, no.~9,
  pp.~1839--1846, 2008.

\bibitem{Kim2022}
J.~Kim, J.~Moon, J.~Ryu, and G.~Lee, ``{CVGC-II}: A new version of a compact
  variable gravity compensator with a wider range of variable torque and
  energy-free variable mechanism,'' {\em IEEE/ASME Transactions on
  Mechatronics}, vol.~27, no.~2, pp.~678--689, 2022.

\bibitem{Mathews2021}
C.~W. Mathews and D.~J. Braun, ``Parallel variable stiffness actuators,'' in
  {\em IEEE/RSJ International Conference on Intelligent Robots and Systems},
  (Prague, Czech Republic), pp.~8225--8231, September 2021.

\bibitem{Plooij2017}
M.~Plooij, W.~Wolfslag, and M.~Wisse, ``{Clutched elastic actuators},'' {\em
  IEEE/ASME Transactions on Mechatronics}, vol.~22, no.~2, pp.~739--750, 2017.

\bibitem{Krimsky2020}
E.~Krimsky and S.~H. Collins, ``Optimal control of an energy-recycling actuator
  for mobile robotics applications,'' in {\em 2020 IEEE International
  Conference on Robotics and Automation (ICRA)}, (Paris, France),
  pp.~3559--3565, May-August 2020.

\bibitem{Kim2021}
S.~Y. Kim and D.~J. Braun, ``Novel variable stiffness spring mechanism:
  Modulating stiffness independent of the energy stored by the spring,'' in
  {\em IEEE International Conference on Intelligent Robots and Systems},
  (Prague, Czech Republic), pp.~8232--8237, September 2021.

\bibitem{Dohring1993}
M.~Dohring, E.~Lee, and W.~Newman, ``A load-dependent transmission friction
  model: theory and experiments,'' in {\em [1993] Proceedings IEEE
  International Conference on Robotics and Automation}, (Atlanta, GA, USA),
  pp.~430--436 vol.3, May 1993.

\bibitem{Kong2012}
K.~Kong, J.~Bae, and M.~Tomizuka, ``A compact rotary series elastic actuator
  for human assistive systems,'' {\em IEEE/ASME Transactions on Mechatronics},
  vol.~17, no.~2, pp.~288--297, 2012.

\bibitem{Braun2019b}
D.~J. Braun, V.~Chalvet, T.-H. Chong, S.~S. Apte, and N.~Hogan, ``Variable
  stiffness spring actuators for low-energy-cost human augmentation,'' {\em
  IEEE Transactions on Robotics}, vol.~35, no.~6, pp.~1435--1449, 2019.

\end{thebibliography}
\end{document}